\documentclass{article}
\usepackage{ijcai17}

\usepackage{times}
\usepackage{latexsym}
\usepackage{mathtools}
\usepackage{array}
\usepackage{amsmath}
\usepackage{comment}
\usepackage{helvet}
\usepackage{courier}
\usepackage{multirow}
\usepackage{graphicx}
\usepackage{epsfig}
\usepackage{algorithm,algpseudocode}
\usepackage{mathdots}
\usepackage{subfloat}
\usepackage{url}
\usepackage{caption}
\usepackage{subcaption}
\usepackage{balance}

\newcommand{\iflong}[1]{}
\newcommand{\moreResult}[1]{}

\newcommand{\ct}[1]{}
\newcommand{\wc}[1]{\textbf{}}
\newcommand{\bc}[1]{\textbf{}}
\newenvironment{ppr}%
  {\begin{minipage}[t]{\textwidth}\begin{tabbing}123\=123\=123\=\kill~\\}%
  {\\\end{tabbing}\end{minipage}}

\title{Using Graphs of Classifiers to Impose Declarative Constraints on Semi-supervised Learning}

\author{Lidong Bing\\
	    AI Lab \\
	    Tencent Inc.\\
	    {\tt lyndonbing@tencent.com}
	  \And
	William W. Cohen\\
	    Machine Learning Department\\
	    Carnegie Mellon Univeristy\\
	    {\tt wcohen@cs.cmu.edu}
	  \And
	  Bhuwan Dhingra\\
	    Language Technologies Institute\\
	    Carnegie Mellon Univeristy\\
	    {\tt bdhingra@cs.cmu.edu}}

\begin{document}

\maketitle

\begin{abstract}
We propose a general approach to modeling semi-supervised learning (SSL) algorithms.  Specifically, we present a declarative language for modeling both traditional supervised classification tasks and many SSL heuristics, including both well-known heuristics such as co-training and novel domain-specific heuristics. In addition to representing individual SSL heuristics, we show that multiple heuristics can be automatically combined using Bayesian optimization methods. We experiment with two classes of tasks, link-based text classification and relation extraction.  We show modest improvements on well-studied link-based classification benchmarks, and state-of-the-art results on relation-extraction tasks for two realistic domains.
\end{abstract}

\section{Introduction}

Most semi-supervised learning (SSL) methods  operate by introducing
``soft constraints'' on how a learned classifier will behave on unlabeled instances, with different constraints leading to different SSL methods.
For example, logistic regression with entropy regularization \cite{grandvalet2004semi}
and transductive SVMs \cite{Joachims:1999:TIT:645528.657646} constrain the classifier to make
confident predictions at unlabeled points, and similarly, many graph-based SSL
approaches require that the instances associated with the endpoints of
an edge have similar labels or embeddings \cite{belkin2006manifold,talukdar2009new,DBLP:series/lncs/WestonRMC12,DBLP:conf/nips/ZhouBLWS03,ZhuICML2003}.  Other
weakly-supervised methods also can be viewed as imposing constraints
on predictions made by a classifier: for instance, in
distantly-supervised information extraction, a useful constraint requires that the classifier, when applied to the set
$S$ of mentions of an entity pair that is a member of relation $r$,
classifies at least one mention in $S$ as a positive instance of $r$
\cite{hoffmann2011knowledge}.

Here we propose a general approach to modeling SSL
constraints.  We will define a succinct
declarative language for specifying semi-supervised learners.  We call our declarative SSL framework the \textit{D-Learner}.

As observed by previous researchers \cite{zhu2005semi}, the appropriate use of SSL is often domain-specific.  The D-Learner framework allows us to define various constraints easily.  Combined with a
tuning strategy with Bayesian Optimization, we can collectively evaluate the
effectiveness of these constraints so as to obtain tailor-made SSL settings for
individual problems.
To examine the efficacy of D-Learner, we apply it to two tasks: link-based text classification and
relation extraction. For the link-based classification task, we will show the flexibility of our declarative language
by defining several SSL constraints for exploiting network structures.
For the relation extraction task,
we will show how our declarative language can express several
intuitive problem-specific SSL constraints.
Comparison against existing SSL methods
\cite{hoffmann2011knowledge,bing-aaai:2017,Surdeanu:2012:MML:2390948.2391003}
shows that D-Learner achieves significant improvements over the state-of-the-art on two domains.


\section{D-Learner: Declaratively Specifying Constraints for Semi-supervised Learners}

\subsection{An Example: Supervised Classification}

\begin{figure}[!t]

\centerline{\small{
\begin{ppr}
predict(X,Y) $\leftarrow$\\
  \>pickLabel(Y) $\wedge$\\
  \>classify(X,Y).\\
classify(X,Y) $\leftarrow$ true\\
  \>$\{$ \textit{f(W,Y): hasFeature(X,W) $\}$}.
\end{ppr}~~\begin{ppr}
mutexFailure(X) $\leftarrow$\\
\>pickMutex(Y1,Y2) $\wedge$\\
\>classify(X,Y1) $\wedge$\\
\>classify(X,Y2).
\end{ppr}}}
\vspace{-3mm}
\caption{Declarative specifications of the models for supervised
  learning, on the left, and for a mutual-exclusivity constraint, on
  the right.}
\label{ppr:sl+mutex}

\smallskip

\centerline{\includegraphics[width=0.5\textwidth]{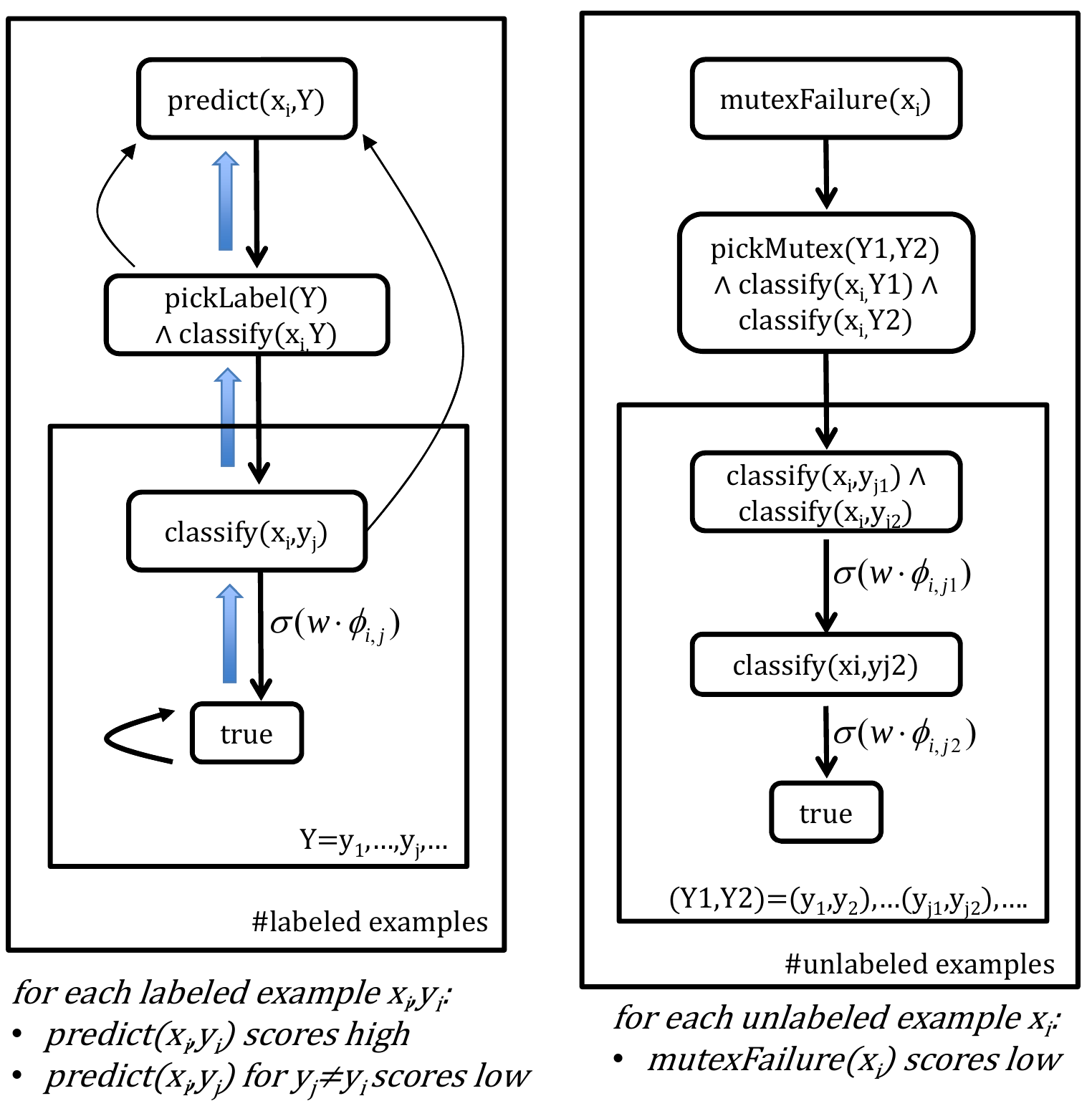}}
\caption{Plate diagrams for supervised learning on the left, and a
  mutual-exclusivity constraint on the right.}
\label{fig:sl+mutex}
\end{figure}

We begin with a simple example.  The left-hand side of
Figure~\ref{ppr:sl+mutex} illustrates how traditional supervised
classification can be expressed declaratively in a logic program. Following the
conventions used in logic programming, capital letters are universally quantified variables.  We extend traditional logic programs by allowing rules to be annotated with a set of
\textit{features}, which will be then weighted to define a strength for the
rule (as we will describe below).   The symbol \textit{true} is a goal that always succeeds. \footnote{We
omit, for brevity, the problem specific definition of
\textit{pickLabel(Y)}, which would consist of rules for each possible
label $y_i$, i.e. \textit{pickLabel($y_1$) $\leftarrow$ true}, \ldots,
\textit{pickLabel($y_N$) $\leftarrow$ true}.}

D-learner programs are associated with a
backward-chaining proof process, can be read
either as logical constraints, or as a non-deterministic program,
which is invoked when a query is submitted.  If the queries processed
by the program are all of the form \textit{predict($x_i$,Y)} where
$x_i$ is a constant, the theory on the left-hand side of
Figure~\ref{ppr:sl+mutex} can be interpreted as saying:
(1) To prove the goal of the form \textit{predict($x_i$,Y)}---i.e.,
  to predict a label $Y$ for the instance $x_i$---non-determin\-istic\-ally pick a possible class label $y_j$, and then prove the goal \textit{classify($x_i$,$y_i$)};
(2) proofs for every the goal \textit{classify($x_i$,$y_j$)}
  immediately succeed, with a strength based on a weighted combination
  of the features in the set $\{ f(w,y_j) : hasFeature(x_i,w) \}$.  This
  set is encoded in the usual way as a sparse vector $\phi_{x_i,y_j}$,
  with one dimension for every object of the form $f(w,y_j)$ where
  $y_j$ is a class label and $w$ is a vocabulary word.  For example,
  if the vocabulary contains the word \textit{hope} and
  \textit{sports} is a possible label, then one feature in
  $\phi_{x_i,y_j}$ might be active exactly when document $x_i$
  contains the word $hope$ and $y_j=\textit{sports}$.
The set of proofs associated with this theory are described by the
plate diagram on the left-hand side of Figure~\ref{fig:sl+mutex}, where the upward-pointing blue arrows denote logical implication, and
the repetition suggested by the plates has the same meaning as in
graphical models.  The black arrows will be explained below.

\subsection{Semantics of D-Learner Constraints}

The D-learner constraints are implemented in a logic programming language called ProPPR \cite{wang2013programming}, whose semantics are defined by a slightly different graph. The ProPPR graph
contains the same set of nodes as the proof graph, but is weighted,
and has a different edge set---namely, the downward-pointing black arrows, which run opposite to the implication edges. For this example, these edges describe
a forest, with one tree for each labeled example $x_i$.
Each tree in the forest begins branching at distance two from the root, and will
have $N$ nodes labeled \textit{true} where $N$ is the number of
possible class labels.
The forest is further augmented with some
additional edges: in particular, we will add a self-loop to each
\textit{true} node, and a ``reset'' edge that returns to the
root (the curved upward-pointing arrows) for each non-\textit{true}
node. (To simplify, the reset and self-loop edges are only shown in the first plate diagram.)
The light upward-pointing edges will have an implicit weight of
$\alpha$, for some fixed $0<\alpha<1$, unannotated edges have an
implicit weight of one, and the weight of feature-annotated edges will
be discussed below.  Finally, if feature vector $\phi$ is associated with a rule, then the
edges produced using that rule are annotated with the feature
vector. (We abbreviate $\phi_{x_i,y_j}$ with $\phi_{i,j}$ in the
figure.)  The weight of such an edge is $\sigma(\mathbf{w}\cdot\phi_{i,j})$, where
$\mathbf{w}$ is a learned parameter vector, and $\sigma$ is a nonlinear
``squashing'' function (e.g., $\sigma(x)\equiv\frac{1}{1+e^{-x}}$).  In
the example, these edges are the ones that exit a node labeled
\textit{classify($x_i$,$y_j$)}.  These are labeled with the function
$\sigma(\mathbf{w}\cdot\phi_{i,j})$, and recall that the vector $\phi_{i,j}$
encodes features associated with the document $x_i$ with the
label $y_j$.

We can now define a Markov process, based on repeatedly transitioning
from a node $v$ to its neighbors, using the normalized outgoing edges
of $v$ as a transition function.  This process is well-defined for any
set of positive edge weights (and positivity can be ensured by the
choice of $\sigma$) and any graph architecture, and will assign a
probability score $\pi_w(v)$ to every node $v$.  Conceptually, it can
be viewed as a probabilistic proof process, with the ``reset''
corresponding to abandoning a proof \cite{wang2013programming}.

Imagine that $\mathbf{w}$ has been trained so that for an example $x_i$ with
true label $y_{j*}$, $\mathbf{w}\cdot\phi_{i,j*}$ has the largest score over
all the possible labels $y_j$.  Note that every \textit{true} node
corresponds to a label $y_j$.  It is not hard to see that the ordering
of $\pi(v)$ over the \textit{true} nodes in the graph for $x_i$
maintains a close correspondence to classification performance.  In
particular, even though the graph is locally normalized, the
$\phi_{i,j}$-labeled edges compete with the upward-pointing
``reset'' edges that lead to the root, so a larger weight will
direct more of the probability mass of the walk toward the
\textit{true} node associated with label $j*$.  More specifically, the
\textit{true} node associated with label $y_{j*}$ will have the
highest $\pi$ of any \textit{true} node in the graph.
Thus, in the example, \textit{there is a close connection between the
  problem of setting $\mathbf{w}$ to minimize empirical loss of the classifier,
  and setting $\mathbf{w}$ to satisfy constraints on the Markov-walk scores of
  the nodes in the graph.}

In ProPPR, it is possible to train weights to maximize or minimize the
score of a particular query response: i.e., one can say that for the
query \textit{predict($x_i,Y$)} responses where $Y=y_{j*}$ are
``positive'' and responses where $Y=y_{j'}$ for $j'\not=j*$ are
``negative''.
Specifically a positive example $a$ incurs a
  loss of $\log_{\sum_{v\in{}V} \pi(v)}$, where $V$ is the set of
  \textit{true} nodes that support answer $a$, and a negative example
  incurs a loss of $\log_{1-\sum_{v\in{}V} \pi(v)}$.
The training data needed for the supervised learning case is indicated in the
bottom of the left-hand of the plate diagram.  Learning is performed
by stochastic gradient descent \cite{wang2013programming}.

\subsection{An example: Semi-supervised Learning}
\label{sec:ssl_mutex}

We finally turn to the right-hand parts of Figures \ref{ppr:sl+mutex}
and \ref{fig:sl+mutex}, which are a D-learner specification for an SSL method.  These rules are a consistency
test to be applied to each \emph{unlabeled} example $x_i$.  In ordinary
classification tasks, any two distinct classes $y_j$ and $y_{j'}$
should be mutually exclusive.  The theory on the right-hand side of
Figure~\ref{ppr:sl+mutex} asserts that a ``mutual exclusion failure''
(mutex\-Failure) occurs if $x_i$ can be classified into two distinct
classes. \footnote{We omit the definition of
\textit{pickMutex(Y1,Y2)}, which would consist of trivial rules for
each possible distinct label pair $y_j$ and $y_{j'}$.}  The corresponding
plate diagram is shown in Figure \ref{fig:sl+mutex}, simplified by
omitting the ``reset'' edges and the self-loops on \textit{true}
nodes.
To (softly) enforce this constraint, we need only to introduce
negative examples for each unlabeled example $x_i$, specifying that
proofs for the goal \textit{mutexFailure($x_i$)} should have low
scores.

Conceptually, this constraint encodes a common bias of SSL systems: the decision boundaries should be drawn in
low-probability regions of the space.
For instance, transductive SVMs maximize the ``unlabeled data margin''
based on the low-density separation assumption that a good
decision hyperplane lies on a sparse area of the feature
space \cite{Joachims:1999:TIT:645528.657646}.
In this case, if a decision
boundary is close to an unlabeled example, then more than one
\textit{classify} goals will succeed with a high score.

\section{Link-based Classification with D-Learner}
\label{sec:Link-based}
The example above explains in detail how to specify one particular type of constraint which is widely used in the past SSL works.  Of course, if this was the only type of constraint that could be specified, D-Learner would not be of great interest: the value of D-Learner is that it allows one to succinctly specify (and implement) many other plausible constraints that can potentially improve learning.
Here we show its application in the task of link-based text classification.

\subsection{The Task and Constraints}

Many real-world datasets contain interlinked entities (e.g. publications linked by citation relation)
and exhibit correlations among labels of linked entities.
Link-based classification aims at improving classification accuracy by exploiting
such link structures besides the attribute values (e.g., text features) of each entity.
In this task, we classify each publication into a pre-defined class,
e.g. Neural\_Networks.
Each publication has features in two views: text content view and citation view.
Thus, each publication can be represented as features of its terms or features of its citations.

Similar to the classifier defined above, we have the text view classifier:\\
\textsf{\small{
\begin{tabular}{l}
predictT(X,Y) $\leftarrow$pickLabel(Y) $\wedge$classifyT(X,Y).  \\
classifyT(X,Y) $\leftarrow$ true ~~ $\{$ \textit{f(W,Y): hasFeature(X,W) $\}$}.
\end{tabular}}}

\noindent And the citation view classifier:\\
\textsf{\small{
\begin{tabular}{l}
predictC(X,Y) $\leftarrow$pickLabel(Y) $\wedge$classifyC(X,Y).\\
classifyC(X,Y)   $\leftarrow$ true ~~ $\{$ \textit{ g(Cited,Y): cites(X,Cited)} $\}$.
\end{tabular}}}

Mutual-exclusivity constraints \textsf{\small{mutexFailureT}} and \textsf{\small{mutexFailureC}}
of the text and citation classifiers are defined in the same way as done in Section \ref{sec:ssl_mutex}.

\indent\textbf{Cotraining constraints.}
D-Learner coordinates the classifiers of two views
to make consistent predictions on testing data by imposing penalty when they disagree:\\
\textsf{\small{
\begin{tabular}{l}
coFailure(X) $\leftarrow$ pickMutex(Y1,Y2) $\wedge$classifyT(X,Y1) $\wedge$ \\
\hspace{20mm}classifyC(X,Y2). \\
coFailure(X) $\leftarrow$ pickMutex(Y1,Y2) $\wedge$classifyC(X,Y1) $\wedge$ \\
\hspace{20mm}classifyT(X,Y2). \\
\end{tabular}}}

\begin{table}[!t]
\caption{Datasets for link-based classification task.\label{t:getoor_data}}
\center{
\begin{small}
\begin{tabular}{l|l|l|l}
\hline
 & CiteSeer & Cora & PubMed \\
\hline
 \# of publications & 3,312 & 2,708  & 19,717   \\
 \# of citation links & 4,732  & 5,429   & 44,338    \\
 \# classes & 6  & 7 & 3    \\
 \# of unique words & 3,703  & 1,433   &  500    \\
\hline
\end{tabular}
\end{small}
}
\end{table}


\indent\textbf{Propagation constraints.}
An initial narrative of label propagation (LP) algorithms is that the neighbors of a good labeled example should be classified consistently with it. To express this in the text view, D-Learner penalizes the violators by: \\
\textsf{\footnotesize{
\begin{tabular}{l}
lpFailure1(X,Y) $\leftarrow$ sim1(X,Z)$\wedge$pickMutex(Y,!Y)$\wedge$predictT(Z,!Y).
\end{tabular}}}

\noindent where pickMutex(Y,!Y) can be replace with pickMutex(!Y,Y) to get another constraint,
and sim1 is defined as:\\
\textsf{\small{
\begin{tabular}{l}
sim1(X1,X2) $\leftarrow$ near(X1,Z)$\wedge$sim1(Z,X2). \\
sim1(X,X) $\leftarrow$ true.
\end{tabular}}}

\noindent If X1 cites or is cited by Z, then near(X1,Z) is true. Thus, lpFailure1 encourages
the publications that have a citation path to X to take the same label as X.

Extending the above one-step walk based sim1 clause, we define a ``two-step'' walk based one:\\
\textsf{\small{
\begin{tabular}{l}
sim2(X1,X2) $\leftarrow$ near(X1,Z1)$\wedge$near(Z1,Z2)$\wedge$sim2(Z2,X2). \\
sim2(X,X) $\leftarrow$ true.
\end{tabular}}}

\noindent Accordingly, the constraint using sim2 is referred to as \textsf{\small{lpFailure2}}. It encourages the publications
cite or are cited by the same publication to have the same label.

\indent\textbf{Regularization constraints.}
\wc{discuss connection to laplacian regularization}
D-Learner can implement the well-studied regularization technique in SSL by smoothing
the behavior of a classifier on unlabeled data.
Specifically, for an unlabeled example, we smooth its label and its neighbors' labels by:\\
\textsf{\small{
\begin{tabular}{l}
smoothFailure(X1) $\leftarrow$ pickMutex(Y1,Y2)$\wedge$\\
\hspace{10mm} classifyT(X1,Y1)$\wedge$near(X1,X2)$\wedge$classifyT(X2,Y2).
\end{tabular}}}

\iflong{
\noindent and a single step of two-step walk as: \\
\textsf{\small{
\begin{tabular}{l}
smoothFailure2(X1) $\leftarrow$ pickMutex(Y1,Y2)$\wedge$classifyT(X1,Y1)$\wedge$\\
\hspace{10mm} near(X1,Z)$\wedge$near(Z,X2)$\wedge$classifyT(X2,Y2).
\end{tabular}}}
}

\subsection{Experiments}

\subsubsection{Settings}
We use three datasets from \cite{sen:aimag08}: CiteSeer, Cora and PubMed, with their
statistical information given in Table \ref{t:getoor_data}. For each dataset, we use 1,000 publications for testing,
and at most 5,000 publications for training. Among the training publications, we randomly pick 20
as labeled examples for each class, and the remaining ones are used as unlabeled.
After the examples are prepared, we employ ProPPR to learn multi-class classifiers with $\alpha=0.1$. The maximum epoch number is
40, and we find the training usually converges in less than 10 epochs.
Note that constraints are combined with equal weights, and we control the effect of different constraints by using different numbers of examples from them (details discussed later).


\subsubsection{Results}
We compare with two supervised learning baselines: SL-SVM and SL-ProPPR, which
only employ the text view features to train classifiers with SVMs and ProPPR, respectively.
For SL-SVM, a linear kernel is used
to train multi-class classifiers.
Another compared baseline is SSL-naive, which simply uses all examples of all constraints in the
semi-supervised learning, regardless their actual effects. D-Learner invokes a tuning strategy
to determine how to use the constraints.

We employ accuracy as the evaluation metric, which is defined
as the ratio between the number of correct predictions and the number of total predictions (i.e. the number of testing examples).
The results are given in Table \ref{t:link_clf_results}.
D-Learner consistently outperforms SL-ProPPR on all three datasets, and the relative improvements are about 1.5\% to 5\%.
Compared with SL-SVM, D-Learner achieves better results on the Cora and PubMed datasets, with
relative improvements about 12\% and 6\%, respectively. For the CiteSeer dataset, D-Learner's performance is comparable to SL-SVM. Although each constraint has its own intuitive interpretation,
SSL-naive does not perform well, particularly poor on CiteSeer and Cora, which suggests that
the appropriate use of SSL is often domain-specific \cite{zhu2005semi}, and needs more careful
assessment on the heuristics.

\wc{should you compare the individual constraints to the D-learner, with the baysian optimization stuff?}

\begin{table}[!t]
\caption{Link-based classification results.\label{t:link_clf_results}}
\center{
\begin{small}
\begin{tabular}{l|l|l|l}
\hline
 & CiteSeer & Cora & PubMed \\
\hline
SL-SVM & \textbf{0.558} & 0.520 & 0.665 \\
SL-ProPPR &  0.528  & 0.551 & 0.688\\
SSL-naive  & 0.109 & 0.334 &  0.672 \\
D-Learner  & 0.551 & \textbf{0.581} &  \textbf{0.699} \\
\hline
\end{tabular}
\end{small}
}
\end{table}

\subsubsection{Tuning with Bayesian Optimization}
\label{sec:text_cat_tuning}
To exploit how to use those domain-specific constraints, D-Learner incorporates a Bayesian optimization based tuning method \cite{NIPS2012_4522}, in which a learning algorithm's generalization
performance is modeled as a sample from a Gaussian process.
The released package, Spearmint\footnote{https://github.com/JasperSnoek/spearmint}, allows one to program a wrapper for the communication with the tuned algorithm, while searching the optimal parameters. Specifically, the wrapper passes the parameters suggested by Spearmint into the learning algorithm, then collects the results from the algorithm for Spearmint to generate a new suggestion.

Instead of adding a weight to control one constraint's effect, we adopt a more straightforward way which tunes the example number of a constraint to use.
Thus, we have 6 parameters: the example numbers of coFailure (\#cF), lpFailure1 (\#lpF1), lpFailure2 (\#lpF2), mutexFailureT (\#mFT), mutexFailureC (\#mFC), and smoothFailure (\#sF),
as listed in Table \ref{t:params_linked_data}.
lpFailure2 is found helpful on all three datasets.
The mutual-exclusivity constraint of text view is helpful on Cora and PubMed datasets.
Cotraining classifiers of both citation view and text content view does not
improve the results, it is because, in the three datesets,
the average citation number of publications is about 1.5 to 2.5, which
is too sparse to train a reliable citation view classifier.

\section{Relation Extraction with D-Learner}
\label{sec:ier}

\subsection{The Task}
We note that in many NLP tasks, there are a plausible number of task-specific constraints which can be easily formulated by a domain expert.
In this section we will describe a number of constraints of
\textit{relation extraction for entity-centric corpora}. Each document in an entity-centric corpus
describes aspects of a particular entity (called \textit{subject or title entity}),
e.g. each Wikipedia article is such a document. Relation extraction
from an entity-centric document is reduced to predicting the relation
between the subject entity (i.e. the first argument of the relation) and an entity mention (i.e. the second argument) in the document.
For example, from a drug article ``Aspirin'', if the target relation is ``sideEffects'',
we need to predict for each candidate (extracted from a single sentence)
whether it is a side effect of ``Aspirin''.
If no such relation holds, we predict the special label ``Other''. This task in medical domain was initially proposed in \cite{bing-aaai:2016,bing-aaai:2017}, where a reader could find more details.

\begin{table}[!t]
\caption{Parameter tuning for link-based classification.\label{t:params_linked_data}}
\center{
\begin{small}
\begin{tabular}{@{~}l@{~}|@{~}c@{~}@{~}c@{~}|@{~}c@{~}@{~}c@{~}|@{~}c@{~}@{~}c@{~}}
\hline
   & \multicolumn{2}{c}{Citeseer}   & \multicolumn{2}{c}{Cora}   & \multicolumn{2}{c}{PubMed} \\
\cline{2-7}
  & Range & Optimal  & Range & Optimal  & Range & Optimal \\
\hline
\#cF &  [0, 2,192] & 0  &  [0, 1,568] & 0  &  [0, 4,940] & 0  \\
\#lpF1  &   [0, 120] & 0  &  [0, 140] & 0  &  [0, 60] & 0  \\
\#lpF2 &   [0, 120] & 80  &  [0, 140] &  140 &  [0, 60] &  60 \\
\#mFT &   [0, 2,192] & 0  &  [0, 1,568] &  1,568 &  [0, 4,940] & 4,940   \\
\#mFC &   [0, 2,192] & 0  &  [0, 1,568] & 0  &  [0, 4,940] & 0 \\
\#sF &   [0, 2,192] & 0  &  [0, 1,568] & 0  &  [0, 4,940] & 0  \\

\hline
\end{tabular}
\end{small}
}
\end{table}

\subsection{Co-training Relation and Type Classifiers}

The second argument of a relation is usually
of a particular unary type. For example, the second argument of the ``interactsWith'' relation is always a drug. Therefore, a plausible idea
is to impose an agreement constraint between relation and type classifiers, analogous to enforcing agreements between classifiers of different views.

With our declarative language, we can coordinate the tasks
of predicting relation and type as follows: \\
\textsf{\small{
\begin{tabular}{l}
coFailure(X)$\leftarrow$ predictR(X,Y)$\wedge$pickRealLabel(Y)$\wedge$\\
\hspace{10mm}inRangeT(Y,T1)$\wedge$pickMutexT(T1,T2)$\wedge$predictT(X,T2),
\end{tabular}}}

\noindent where \textit{pickRealLabel(Y)} consists of trivial rules for each
non-``Other'' relation label, and \textit{inRangeT(Y,T1)} consists of trivial rules
for each relation label and its value range type, \textit{pickMutexT(T1,T2)}
consists of trivial rules of each distinct pair of type labels,
\textit{predictR} and \textit{predictT} are relation and type classifiers, respectively.
In other words, \textit{coFailure($x_i$)} says we should penalize cases where
a mention $x_i$ is predicted having the real relation label Y
(i.e. not ``Other''), whose range is type T1,
but the predicted type of $x_i$ is T2 which is mutually exclusive with T1.
Thus, by specifying that proofs for the goal \textit{coFailure($x_i$)} having low
scores, classifiers that make this sort of error will be downweighted.

Mutual-exclusivity constraints \textsf{\small{mutexFailureT}} and \textsf{\small{mutexFailureR}}
of type and relation classifiers are defined in the same way as we did before.

\iflong{
Similar to the mutexFailure constraint for the relation classifier, we have the below for
the type classifier: \\
\textsf{\small{
\begin{tabular}{l}
mutexFailureT(X) $\leftarrow$ pickMutexT(Y1,Y2) $\wedge$\\
\hspace{20mm}classifyT(X,Y1) $\wedge$classifyT(X,Y2).
\end{tabular}}}

\subsubsection{More Type-based Constraints}

Noun phrases (NPs) in a coordinate-term list (e.g., the underlined ones in ``Get medical help if you experience \underline{chest pain}, \underline{weakness}, or \underline{shortness of breath}'') are likely to have the same category, a fact used in a prior work~\cite{bing-EtAl:2015:EMNLP}.
To explore this intuition, we define a constraint:
\centerline{
\begin{ppr}
nearTTL(X1,X2) :- predT(X1,Y),classifyT(X2,Y)
\end{ppr}}}

\subsection{SSL Constraints for Relation Extraction}

\textbf{Document constraint.} If an entity string appears as multiple mentions
in a document, they should have the same relation label (relative to the subject), or some have ``Other'' label: \\
\textsf{\small{
\begin{tabular}{l}
docFailure(X1,X2)  $\leftarrow$ pickMutex(Y1,Y2)$\wedge$pickRealLabel(Y1)$\wedge$\\
\hspace{10mm}pickRealLabel(Y2)$\wedge$classify(X1,Y1)$\wedge$classify(X2,Y2).
\end{tabular}}}

\noindent For example, if ``heartburn'' appears twice in the ``Aspirin'' document,
we should make consistent predictions, i.e. ``sideEffects'' for them, or
predict one or both as ``Other''.
Note that a mention can also refer to a coordinate-term list, such as
``vomiting, headache and nausea'', of multiple NPs. In fact, we compute
the intersection of the NP sets of two mentions, and if it is not empty,
this constraint applies, such as for ``vomiting, and headache''
and ``vomiting, and heartburn''.

\textbf{Sentence constraint}. For each sentence, we require that only one
mention can be labeled as a particular relation
(this requirement is usually satisfied in real data): \\
\textsf{\small{
\begin{tabular}{l}
sentFailure(X1,X2)  $\leftarrow$ pickRealLabel(Y1)$\wedge$pickRealLabel(Y2)$\wedge$\\
\hspace{20mm} classify(X1,Y1)$\wedge$classify(X2,Y2),
\end{tabular}}}

\noindent where X1 and X2 are a pair of mentions extracted from a single sentence.
This constraint penalizes extracting multiple relation objects from a single sentence.
For example, from the sentence ``Some products that may interact with this drug include: Aliskiren and ACE inhibitors'', ``some products'' and ``Aliskiren and ACE inhibitors'' should not be labeled
as the interactsWith relation simultaneously.

\textbf{Section title constraint.} In some entity-centric corpora, the content of
a document is organized into sections. This constraint basically
says, for two mentions of the same NP in the same
section (currently determined simply by matching
section titles) of two documents, they should have the same
relation label, relative to their own document subjects:\\
\textsf{\small{
\begin{tabular}{l}
titleFailure(X1,X2)   $\leftarrow$ pickMutex(Y1,Y2)$\wedge$pickRealLabel(Y1)$\wedge$\\
\hspace{10mm} pickRealLabel(Y2)$\wedge$classify(X1,Y1)$\wedge$classify(X2,Y2).
\end{tabular}}}

\noindent E.g., if ``heartburn'' appears in ``Adverse reactions''
sections of drugs ``Aspirin'' and ``Singulair'', it is plausible
to infer both mentions are ``sideEffects'', or one or both are ``Other''.

\subsection{Experiments}

\subsubsection{Settings}

We follow the experimental settings in \cite{bing-aaai:2017}. The drug corpus is DailyMed containing 28,590 articles, and the disease corpus is WikiDisease containing 8,596 articles. 3 and 5 relations are extracted for drugs and diseases, respectively. A pipeline's task then is to extract all correct values of the second argument of a given relation from a test document. The range types of the second arguments of relations are also defined in their paper. For details about preprocessing, features, distant labeling seeds, annoated evaluation and tuning datasets, please refer to \cite{bing-aaai:2017}.

\subsubsection{Training Classifiers}
\label{sec:training_data}
Without manually labeled training relation examples, we employ the seed triples from Freebase to distantly label articles in the corpora, as done in \cite{bing-aaai:2017}.
For instance, the triple sideEffects(Aspirin,heartburn) will label
a mention ``heartburn'' from the Aspirin article as an example
of sideEffects relation.
The raw data by distant labeling is very noisy \cite{riedel2010modeling,bing-EtAl:2015:EMNLP,bing-aaai:2016,bing-aaai:2017},
so we employ the distillation method with LP in \cite{bing-aaai:2017} to get a cleaner set of relation examples, where the top K (the tuning for K will be described later) most confident, scored by LP, examples are used as training data.
We also pick a set of mentions from the testing corpus that
are not labeled as any relation as general negative examples.

The training type examples are collected by DIEL~\cite{bing-EtAl:2015:EMNLP},
which extracts the instances of those range types from Freebase as seeds, and extends
the seed set by LP.
After that, we pick top K as training examples. Similarly, we also randomly pick
a number of general negative type examples.


\subsubsection{Compared Baselines}

We  compare against three existing methods: \textit{MultiR}, \cite{hoffmann2011knowledge} which models each relation mention separately and aggregates their labels using a deterministic OR; \textit{Mintz++} from \cite{Surdeanu:2012:MML:2390948.2391003}, which improves on the original model from \cite{mintz2009distant} by training multiple classifiers, and allowing multiple labels per entity pair; and \textit{MIML-RE} \cite{Surdeanu:2012:MML:2390948.2391003} which has a similar structure to \textit{MultiR}, but uses a classifier to aggregate the mention level predictions into an entity pair prediction. We used the publicly available code from the authors \footnote{http://aiweb.cs.washington.edu/ai/raphaelh/mr/ and http://nlp.stanford.edu/software/mimlre.shtml} for the experiments. We tuned their parameters, including negative example number, epoch number (for both \textit{MultiR} and \textit{MIML-RE}), and training fold number (for \textit{MIML-RE}), directly on the evaluation data and report their best performance.

We have 4 supervised learning baselines via distant supervision:
DS-SVM, DS-ProPPR, DS-Dist-SVM, and DS-Dist-ProPPR.
The first two use the distantly-labeled examples from the testing corpora as training data, while the last two use the distilled examples as training data, as done for D-Learner.
For DS-ProPPR and DS-Dist-ProPPR, ProPPR is used to learn multi-class classifiers, as done for D-Learner.
For DS-SVM and DS-Dist-SVM, binary classifiers are trained with SVMs, and DS-Dist-SVM is exactly the \textit{DIEJOB\_Target} in \cite{bing-aaai:2017}. \footnote{We do not compare with \textit{DIEJOB\_Both}, since it uses some training examples of good quality from an additional corpus.}

\subsubsection{Results}
We evaluate the performance of different systems
from an IR perspective: a title entity (i.e., document name)
and a relation together act as a query, and the extracted NPs as retrieval results.
The evaluation results are given in Table \ref{t:prf}.
The systems with ``*'' are directly tuned on the evaluation data.
Other systems are tuned with a tuning dataset.
For the disease domain, D-Learner achieves its best performance without using any examples from the
coFailure constraint, while for the drug domain, it achieves its best performance without using any
examples from docFailure, secFailure, and titleFailure constraints.
The tuning details will be discussed later.

\begin{table}[!t]
\center{
\caption{Relation extraction results.\label{t:prf}}
\begin{small}
\begin{tabular}{l|@{~}c@{~}@{~}c@{~}@{~}c@{~}|@{~}c@{~}@{~}c@{~}@{~}c}
  \hline
 &  & Disease &  &  & Drug & \\
\cline{2-7}
& P &    R   & F1 & P & R & F1 \\
\hline
DS-SVM & 0.228 & 0.335 & 0.271 & 0.170 & 0.188 & 0.178 \\
DS-ProPPR & 0.159 & 0.354 & 0.219 & 0.158 & 0.188 & 0.172 \\
DS-Dist-SVM & 0.231 & 0.337 & 0.275 & 0.299 & 0.300 & 0.300 \\
DS-Dist-ProPPR & 0.184 & 0.355 & 0.243 & 0.196 & 0.358 & 0.253 \\
\hline
MultiR* & 0.198 & 0.333 & 0.249 & 0.156 & 0.138 & 0.146  \\
Mintz++* & 0.192 & 0.353 & 0.249 & 0.177 & 0.178 & 0.178  \\
MIML-RE* & 0.211 & \textbf{0.360} & 0.266 & 0.167 & 0.160 & 0.163  \\
\hline
D-Learner & \textbf{0.378} & 0.271 & \textbf{0.316} & \textbf{0.403} & \textbf{0.356} & \textbf{0.378} \\

\hline
\end{tabular}
\end{small}
}
\end{table}

\begin{figure*}[!t]
\begin{subfigure}{.5\textwidth}
  \centering
  \includegraphics[width=\linewidth]{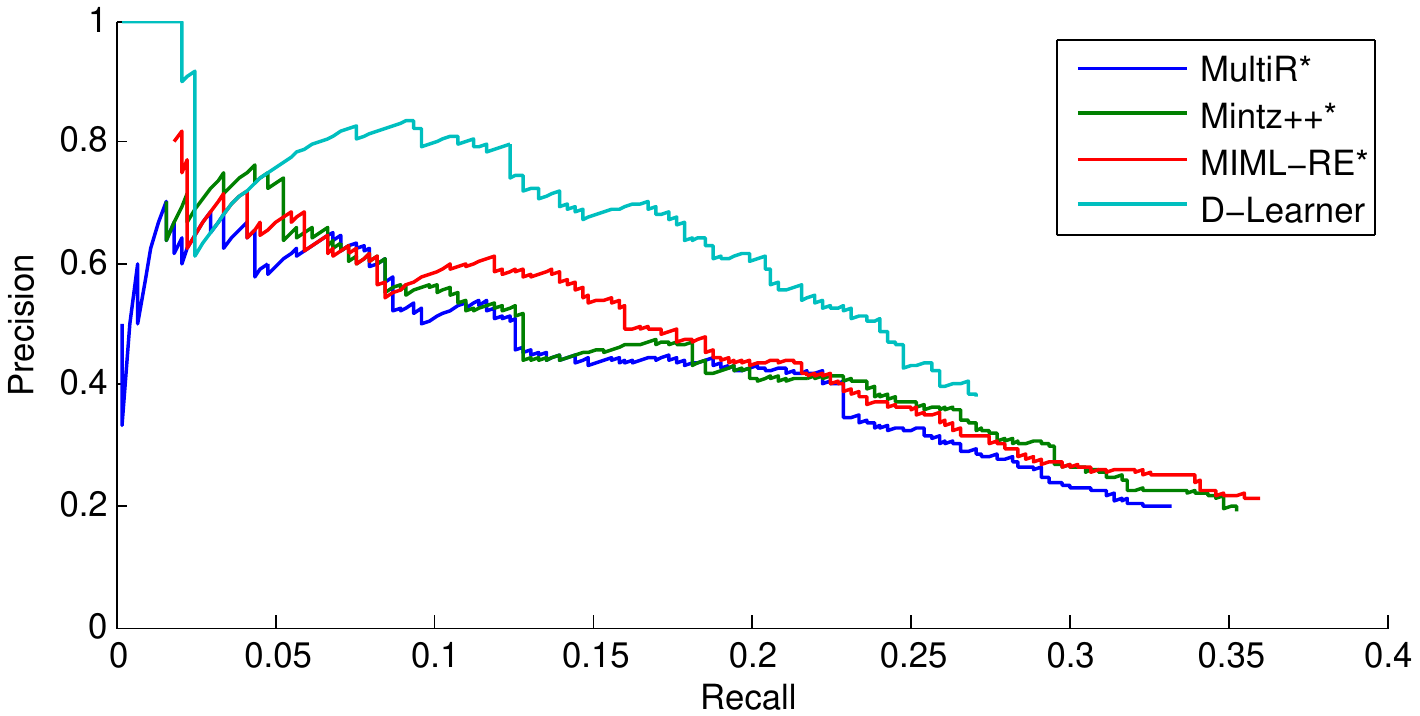}
  \caption{Disease domain.}
  \label{fig:disease_pr}
\end{subfigure}%
\begin{subfigure}{.5\textwidth}
  \centering
  \includegraphics[width=\linewidth]{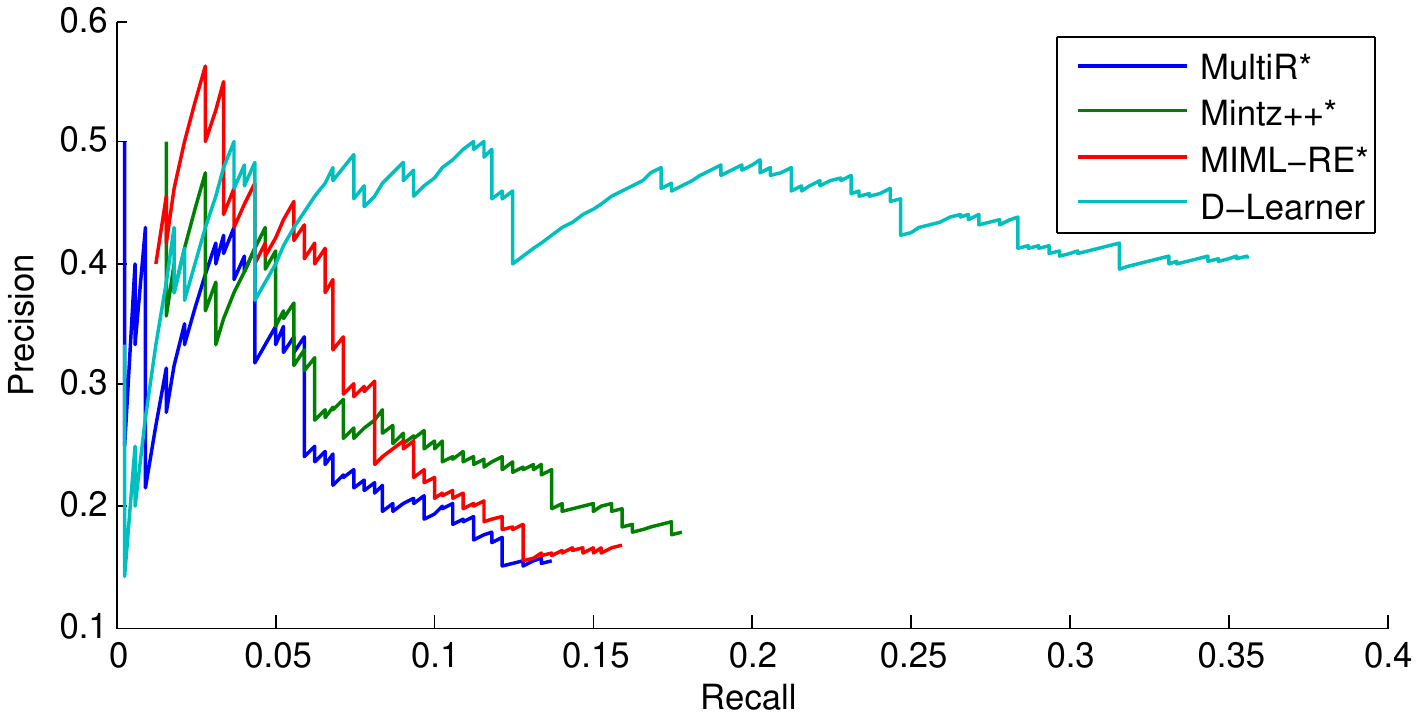}
  \caption{Drug domain.}
  \label{fig:drug_pr}
\end{subfigure}
\caption{\label{fig:pr_curve}Precision-recall curves.}
\end{figure*}

D-Learner outperforms all the other systems. Compared with MultiR,
Mintz++, and MIML-RE, the relative improvements under F1 are 19\% to 27\% in the disease domain,
and 112\% to 159\% in the drug domain.
This result shows that our D-Learner framework has overall superiority on
the relation extraction task, mainly because of its capability of specifying
these tailor-made constraints.
The precision values of D-Learner are much higher than all compared systems on both domains.
For recall, D-Learner performs much better than MultiR,
Mintz++, and MIML-RE on the drug domain, and not as good as
them on the disease domain.

The SVM-based baselines outperform the ProPPR-based baselines, which shows that the good performance of D-Learner
on this task is not because of using ProPPR for its implementation.
Comparisons between DS-X and DS-Dist-X show that the distillation step is useful for
better results.

Precision-recall curves are given in Figures \ref{fig:disease_pr} and \ref{fig:drug_pr}.
For the disease domain, D-Learner's precision is almost always better,
at the same recall level, than any of the other methods. For the drug domain,
D-Learner's precision is generally better after the recall level 0.05.


\subsubsection{Tuning with Bayesian Optimization}
\label{sec:variants}
As listed in Table \ref{t:params_tuning}, we have 10 parameters: the number of training relation examples (\#R), picked from the top
of the ranked (by LP-based distillation) distantly labeled examples; the number of general negative
relation examples (\#NR); the numbers of type examples (\#T) and general negative type examples (\#NT); the example numbers of coFailure (\#cF), mutexFailureR (\#mFR), mutexFailureT (\#mFT),  docFailure (\#dF), titleFailure (\#tF), and sentFailure (\#sF). The unlabeled constraint examples are randomly picked from the entire testing corpus. The 2nd and the 5th columns give the value ranges, and the 3rd and 6th columns give the step sizes while searching the optimal.
Ten pages are annotated as the tuning data.
We optimize the performance of D-Learner under F1, and the obtained optimal
parameters are given in the 4th and 7th columns.

For both domains, D-Learner achieves better results by
using a portion of top-ranked relation examples as training data.
(This observation is consistent with the comparisons of DS-Dist-ProPPR
vs. DS-ProPPR, and DS-Dist-SVM vs. DS-SVM.)
It shows that using a smaller but cleaner set of training examples, combined with
some SSL constraints and cotraining, can improve the performance.

For the disease domain, D-Learner performs better without cotraining type classifiers,
thus, the optimal type related parameters, i.e. \#T, \#NT, \#cF, and \#mFT, for disease are 0. While for the drug domain,
cotraining can bootstrap the performance of relation extraction.
One explanation is that for a domain, e.g. drug, in which the
second argument values of relations are less
ambiguous (e.g. symptom entity for ``sideEffect''), the cotrained type
classifiers are accurate so that they can help the
relation classifier learner explore useful features that are
not captured by relation training examples.
For the disease domain, the second argument values of relations have more ambiguity
(e.g. ``age'' and ``gender'' for the relation ``riskFactors''),
which causes the learnt type classifiers inaccurate and unhelpful for the relation classifiers.
With respect to the tailor-made constraints for relation extraction,
i.e., docFailure, sentFailure, and titleFailure, D-Learner also
behaves differently on the two domains. For the drug domain, it is optimal
not using any examples from them, but for the disease domain, using all examples
from them is preferred.

\begin{table}[!t]
\center{
\caption{Parameter tuning for relation extraction.\label{t:params_tuning}}
\begin{small}
\begin{tabular}{l|@{~}c@{~}@{~}c@{~}@{~}c@{~}|@{~}c@{~}@{~}c@{~}@{~}c}
  \hline
 &  & Disease &  &  & Drug & \\
\cline{2-7}
 & Range & Step & Optimal & Range & Step & Optimal \\
\hline
 \#R & (0, 5k] & 10 & 2.1k  & (0, 5k] & 10 & 580 \\
 \#NR & [0, 10k] & 100 & 5.5k & [0, 10k] & 100 & 1.2k \\
 \#T & [0, 10k] & 200  & 0  & [0, 10k] & 200 & 10k \\
 \#NT & [0, 20k] & 1k  & 0  & [0, 20k] & 1k & 20k \\
 \#cF & [0, 100k] & 10k  & 0  & [0, 100k] & 10k & 100k \\
 \#mFR & [0, 20k] & 2k & 20k  & [0, 20k] & 2k & 20k \\
 \#mFT & [0, 20k] & 2k & 0  & [0, 20k] & 2k & 20k \\
 \#dF & [0, 30k] & 2k & 30k  & [0, 20k] & 2k & 0  \\
 \#sF & [0, 80k] & 2k & 80k  & [0, 20k] & 2k & 0  \\
 \#tF & [0, 30k] & 2k & 30k  & [0, 20k] & 2k & 0   \\
\hline
\end{tabular}
\end{small}
}
\end{table}

Note that the above conclusions on the effect of relation extraction constraints and
cotraining type classifiers are under the integral setting of D-Learner,
i.e. using both constraints and cotraining simultaneously.
More tuning experiments show that for the drug domain, only using those
relation extraction constraints also improves the performance, however, the
improvement is not as significant as only using the cotraining setting;
For the disease domain, only using the cotraining setting
also improves the results, but not as much as only using those constraints.
While using both constraints and cotraining in the integral setting,
it does not necessarily end up with a value-added effect.

For both domains, the mutexFailure examples are found always helpful.
Note that when $\textnormal{\#T}=0$, the mutexFailureT
constraint becomes meaningless, thus, $\textnormal{\#mFT}=0$ for the disease domain.
Another point to mention is that, as shown in Table \ref{t:params_tuning},
the optimal values of some parameters are the maximums in their ranges, which
indicates that D-Learner is likely to achieve even better performance if more
examples of these constraints are used.





\section{Conclusions and Future Work}
We proposed a general approach, D-Learner, to modeling SSL constraints.
It can approximate traditional supervised learning and many natural SSL
heuristics by declaratively specifying the desired behaviors of classifiers.
With a Bayesian Optimization based tuning strategy, we can collectively evaluate the
effectiveness of these constraints so as to obtain tailor-made
SSL settings for individual problems.
The efficacy of this approach is examined in two tasks,
link-based text classification and relation extraction,
and encouraging improvements are achieved.
There are a few open questions
to explore: adding hyperparameters (e.g., different weights for different constraints); adding more control over the supervised loss versus the constraint-based loss; and testing the approach on other tasks.


\bibliographystyle{ijcai17}

\end{document}